%% file: main.tex
\documentclass[10pt,twocolumn,letterpaper]{article}

\usepackage{cvpr}              

\usepackage{multirow}


\definecolor{cvprblue}{rgb}{0.21,0.49,0.74}
\usepackage[pagebackref,breaklinks,colorlinks,allcolors=cvprblue]{hyperref}

\def\paperID{5864} 
\def\confName{CVPR}
\def\confYear{2025}

\title{DORNet: A Degradation Oriented and Regularized Network for \\Blind Depth Super-Resolution}

\author{Zhengxue Wang$^{1}$\thanks{Equal contribution}, Zhiqiang Yan$^{1*\dagger}$, Jinshan Pan$^{1}$, Guangwei Gao$^{2}$, Kai Zhang$^{3}$, and Jian Yang$^{1}$\thanks{Corresponding authors}\\
$^1$PCA Lab\thanks{PCA Lab, Key Lab of Intelligent Perception and Systems for High-Dimensional Information of Ministry of Education, School of Computer Science and Engineering, Nanjing University of Science and Technology.}\ , Nanjing University of Science and Technology\\
$^2$Nanjing University of Posts and Telecommunications \ \ 
$^3$Nanjing University\\
{\tt\small \{zxwang, yanzq, jspan, csjyang\}@njust.edu.cn}, 
{\tt\small csggao@gmail.com, kaizhang@nju.edu.cn}
}

\begin{document}
\maketitle
\input{sec/0_abstract}

\input{sec/1_intro}

{
    \small
    \bibliographystyle{ieeenat_fullname}
    \bibliography{main}
}


\end{document}

%% file: sec/0_abstract.tex
\begin{abstract}
Recent RGB-guided depth super-resolution methods have achieved impressive performance under the assumption of fixed and known degradation (e.g., bicubic downsampling). However, in real-world scenarios, captured depth data often suffer from unconventional and unknown degradation due to sensor limitations and complex imaging environments (e.g., low reflective surfaces, varying illumination). Consequently, the performance of these methods significantly declines when real-world degradation deviate from their assumptions. In this paper, we propose the Degradation Oriented and Regularized Network (DORNet), a novel framework designed to adaptively address unknown degradation in real-world scenes through implicit degradation representations. Our approach begins with the development of a self-supervised degradation learning strategy, which models the degradation representations of low-resolution depth data using routing selection-based degradation regularization. To facilitate effective RGB-D fusion, we further introduce a degradation-oriented feature transformation module that selectively propagates RGB content into the depth data based on the learned degradation priors. Extensive experimental results on both real and synthetic datasets demonstrate the superiority of our \href{https://github.com/yanzq95/DORNet}{DORNet} in handling unknown degradation, outperforming existing methods. 

\end{abstract}

%% file: sec/1_intro.tex
\section{Introduction}
\label{sec:intro}

\begin{figure}[t]
\centering
\includegraphics[width=0.98\columnwidth]{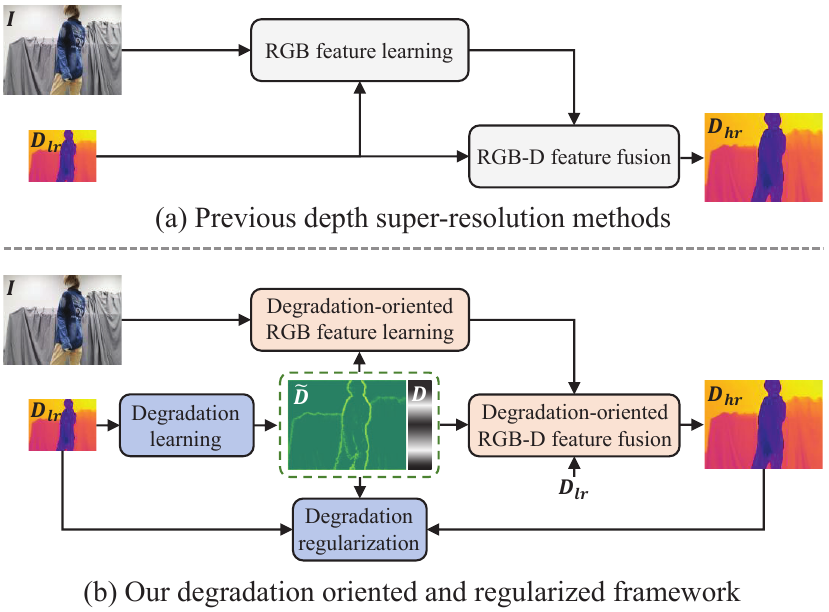}\\
\vspace{-3pt}
\caption{Previous methods (a) directly fuse the RGB information aligned with the LR depth, while our method (b) focuses more on modeling the degradation representation of the LR depth to provide targeted guidance for HR depth recovery.}\label{fig:fram}
\vspace{-6pt}
\end{figure}

Blind depth super-resolution (DSR) aims to recover precise high-resolution (HR) depth from low-resolution (LR) depth with unknown degradation, which has been widely applied in many fields, such as virtual reality~\cite{li2020unsupervised, yan2023distortion, wang2023rgb, wang2024dcdepth}, augmented reality~\cite{yan2022multi, sun2021learning, fan2024depth, yang2024depth, yan2022rignet}, and 3D reconstruction~\cite{wang2021regularizing, choe2021volumefusion, de2022learning, yan2024tri, yang2022codon}. 
As shown in Fig.~\ref{fig:fram}(a), recent RGB-guided DSR methods~\cite{yuan2023recurrent, zhong2021high, yan2023desnet, shin2023task, chen2024intrinsic} have been proposed that integrate RGB features aligned with input depth based on the assumption of known and fixed degradation, achieving excellent performance on synthetic data.
%

However, due to limitations in sensor technology and imaging environments, depth data captured from real-world scenes often suffer from unconventional and unknown degradation~\cite{yan2024completion} (\textit{e.g.}, structural distortion and blur). Such real-world degradation results in structure inconsistency between depth and RGB, impairing the performance of previous methods. 
Moreover, real-world degradation labels are unavailable, preventing us from explicitly estimating the degradation between LR depth and HR depth.

\begin{figure*}[t]
\centering
\includegraphics[width=1.92\columnwidth]{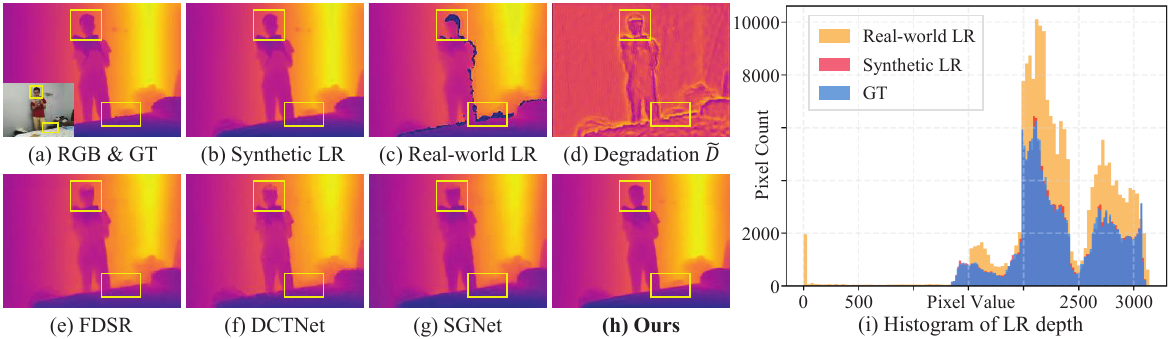}\\
\vspace{-3pt}
\caption{Visual results of LR depth, HR depth, and degradation representation. (b) and (c) are the synthetic and the real-world LR depth, respectively. (d) is the learned degradation representation $\boldsymbol {\tilde{D}}$. (e)-(g) are the HR depth predicted by FDSR~\cite{he2021towards}, DCTNet~\cite{zhao2022discrete}, and SGNet~\cite{wang2024sgnet}, while (h) is produced by our DORNet. (i) is the histogram of real-world LR, synthetic LR, and ground-truth (GT) depth.}\label{fig:RealSecLR}
\vspace{-5pt}
\end{figure*}

As illustrated in Fig.~\ref{fig:RealSecLR}(b) and Fig.~\ref{fig:RealSecLR}(c), the LR depth synthesized using bicubic downsampling exhibits accurate depth structures, while the real-world LR depth experiences more severe structural distortion. Furthermore,  Fig.~\ref{fig:RealSecLR}(i) indicates that the distribution of the real-world LR depth shows a greater difference from the ground-truth depth compared to the synthetic LR depth. This makes it more challenging for DSR to restore accurate HR depth from LR depth with unknown degradation.

To address these issues, as shown in Fig.~\ref{fig:fram}(b), we propose a degradation oriented and regularized network (DORNet). The DORNet utilizes degradation representations to guide the restoration of HR depth from real-world scenarios with unknown degradation. 
To this end, we present a self-supervised degradation learning strategy to estimate the implicit degradation representations between LR and HR depth. In this strategy, a router mechanism is first introduced to dynamically control the generation of degradation kernels with varying scales. 
We then design degradation regularization that leverages these kernels to deteriorate the predicted HR depth, yielding a new degraded depth. Consequently, the entire degradation process is learned by narrowing the distance between the new degraded depth and the LR depth, without using degradation labels. 
Furthermore, we observe that RGB can provide sharp and complete details for the degradation areas of the LR depth. Therefore, we propose utilizing the estimated degradation to adaptively select RGB features to guide and facilitate the RGB-D fusion. Concretely, we develop a degradation-oriented fusion scheme, deploying a degradation-oriented feature transformation module (DOFT). The DOFT produces filter kernels from learned degradation and then filters the RGB features, offering complementary contents for the depth features.

Owing to these designs, Fig.~\ref{fig:RealSecLR}(d) demonstrates that the real-world degradation learned by DORNet accurately characterizes the degradation areas of the LR depth, thereby providing precise guidance for RGB-D fusion. Moreover, compared to previous approaches~\cite{he2021towards, zhao2022discrete, wang2024sgnet}, Fig.~\ref{fig:RealSecLR}(h) reveals that our method can effectively restore HR depth with more accurate and clearer structures. 

In short, our contributions are as follows:
\begin{itemize}
    \item We introduce a novel DSR framework termed DORNet, which utilizes degradation representations to adaptively address unknown degradation in real-world scenes. 
    \item We design a self-supervised degradation learning strategy to model degradation representations of LR depth using routing selection-based degradation regularization. 
    \item We propose a degradation-oriented fusion scheme that selectively transfers RGB content into depth by performing DOFT based on learned degradation priors. 
    \item Extensive experiments demonstrate that our DORNet achieves state-of-the-art performance.
\end{itemize}

\begin{figure*}[t]
\centering
\includegraphics[width=1.92\columnwidth]{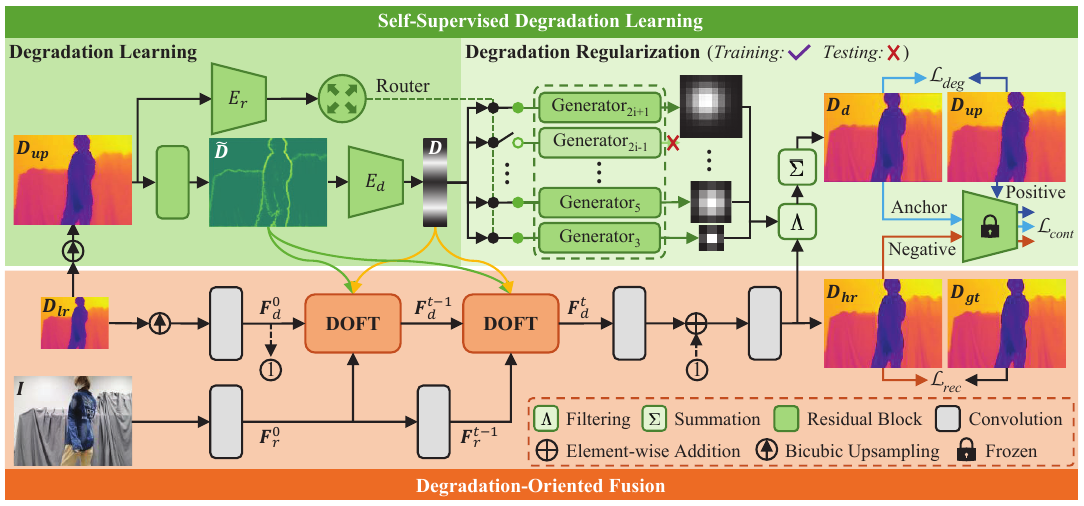}\\
\vspace{-3pt}
\caption{Overview of DORNet. Given $\boldsymbol D_{up} $ as input, the degradation learning first encodes it to produce degradation representations $\boldsymbol {\tilde{D}}$  and $\boldsymbol D $. Then, $\boldsymbol {\tilde{D}}$,  $\boldsymbol D $, $\boldsymbol D_{lr} $, and $\boldsymbol I $ are fed into multiple degradation-oriented feature transformation (DOFT) modules, generating the HR depth $\boldsymbol D_{hr} $. Finally, $\boldsymbol D $ and $\boldsymbol D_{hr} $ are sent to the degradation regularization to obtain $\boldsymbol D_{d} $, which is used as input for the degradation loss $\mathcal{L} _{deg} $ and the contrastive loss $ \mathcal{L}_{cont} $. The degradation regularization only applies during training and adds no extra overhead in testing.}\label{fig:pipeline}
\vspace{-3pt}
\end{figure*}

\section{Related Work}
\label{sec:releatedwork}

\subsection{Depth Map Super-Resolution}

\noindent{\textbf{Synthetic Depth Super-Resolution.}} Many DSR methods~\cite{guo2018hierarchical, tang2021bridgenet, wang2024scene, metzger2023guided} have made significant progress on synthetic data with known degradation. For example, Hui et al.~\cite{hui2016depth} develop a multi-scale guidance network to enhance the boundary clarity of depth. In~\cite{ye2020pmbanet}, Ye et al. utilize the progressive multi-branch fusion network to restore HR depth with sharp boundaries. Recently, a few guided image filtering methods~\cite{li2019joint, kim2021deformable, zhong2023deep} have been proposed for transferring guidance information to the target. For instance, Li et al.~\cite{li2016deep} design a learning-based joint filtering method that propagates salient structures from guidance into target. Kim et al.~\cite{kim2021deformable} apply the deformable kernel network to learn sparse and spatially-variant filter kernels. Additionally, to extract common features from different modality inputs, Deng et al.~\cite{deng2020deep} present a common and unique information splitting network based on multi-modal convolutional sparse coding. Similarly, Zhao et al. build the discrete cosine transform network~\cite{zhao2022discrete} and the spherical spatial feature decomposition network~\cite{zhao2023spherical} to separate the private and shared features between RGB and depth. Unlike these approaches, we focus on utilizing the degradation representations of LR depth to adaptively address unconventional and unknown degradation in real-world scenarios.

\noindent{\textbf{Real-world Depth Super-Resolution.}} Recently, real-world DSR~\cite{liu2018depth, song2020channel, he2021towards, gu2020coupled} targeting unknown degradation has attracted broad attention. For instance, Liu et al.~\cite{liu2016robust} propose a robust optimization framework to address the issues of inconsistency in RGB edges and discontinuity in depth. Song et al.~\cite{song2020channel} employ both non-linear degradation with noise and interval down-sampling degradation to simulate LR depth for real-world DSR. Besides, He et al.~\cite{he2021towards} construct a real-world RGB-D dataset, and design a fast DSR network based on octave convolution. More recently, Yan et al.~\cite{yan2022learning} introduce an auxiliary depth completion branch to recover dense HR depth from incomplete LR depth. Yuan et al.~\cite{yuan2023structure} develop a structure flow-guided model for real-world DSR, which learns a cross-modal flow map to guide the transfer of RGB structural information. Different from previous researches, we pay more attention to modeling the implicit degradation representations of LR depth, and selectively propagating RGB information into depth data based on the estimated degradation priors.

\subsection{Degradation Representation Learning}
Degradation representations have been widely applied in several single-modal image restoration tasks~\cite{wang2021unsupervised, liang2022efficient, zhang2023ingredient}. For example, Wang et al.~\cite{wang2021unsupervised} learn degradation representations for blind image super-resolution by assuming that the degradation of different patches within each image is the same. Similarly, Xia et al.~\cite{xia2023knowledge} develop a degradation estimator based on knowledge distillation to model the degradation representations. Li et al.~\cite{li2022learning}  introduce a multi-scale degradation injection network to jointly optimize reblurring and deblurring. Additionally, some approaches~\cite{li2022all, yin2022conditional, zhang2023ingredient} explore solutions that can be applied to various degradation in a single model. For instance, Li et al.~\cite{li2022all} design an all-in-one image restoration framework, which can recover images with different degradation in one network. Inspired by them, we develop a self-supervised degradation learning strategy to estimate the degradation representations of LR depth using routing selection-based degradation regularization. The learned degradation priors are employed to guide the feature transformation between multi-modal inputs.

\section{Method}
\label{sec:method}

\subsection{Network Architecture}
Given LR depth $\boldsymbol D_{lr}\in R^{h\times w\times 1} $ with unknown degradation and RGB $\boldsymbol I\in R^{sh\times sw\times 3} $ as inputs, our method aims to recover accurate HR depth $\boldsymbol D_{hr}\in R^{sh\times sw\times 1} $ by learning the degradation representations. $h$, $w$, and $s$ represent the height, width, and upsampling factor, respectively.

As shown in Fig.~\ref{fig:pipeline}, our DORNet mainly consists of a self-supervised degradation learning strategy (green part) and a degradation-oriented fusion scheme (orange part). Specifically, the upsampled LR depth $\boldsymbol D_{up}\in R^{sh\times sw\times 1} $ is first input into the degradation learning, producing both the router and the degradation representations, $\boldsymbol {\tilde{D}}$  and $\boldsymbol D $. 
Then, $\boldsymbol {\tilde{D}}$, $\boldsymbol D $, $\boldsymbol D_{lr} $, and $\boldsymbol I$ are sent to multiple degradation-oriented feature transformation modules (DOFT), which selectively propagate RGB information into the depth features, resulting in HR depth $\boldsymbol D_{hr} $. 
Next, the degradation regularization takes $\boldsymbol D $ as input and utilizes routing selection to adaptively generate degradation kernels with varying scales, all of which are sent into the filtering and summation modules together with $\boldsymbol D_{hr} $, obtaining the new degraded depth $\boldsymbol D_{d} $. Finally, $\boldsymbol D_{d} $ is employed as input for the degradation loss $\mathcal{L} _{deg} $ and the contrastive loss~\cite{wu2021contrastive} $ \mathcal{L}_{cont} $, further promoting the learning of degradation representations.

Furthermore, to balance computational complexity and performance, we present a more lightweight DSR model, DORNet-T, which is achieved by reducing all convolutional channels to $\frac{3}{8}$ of those in DORNet, while maintaining the entire network architecture unchanged.

\subsection{Self-Supervised Degradation Learning}

\noindent{\textbf{Degradation Learning.}} As illustrated in Fig.~\ref{fig:pipeline} (upper left), given $\boldsymbol D_{lr} $ as input, bicubic upsampling is first utilized to generate the upsampled depth $\boldsymbol D_{up}$. Then, we employ the residual block $f_{rb}$ and the degradation encoder $E_{d} $ to encode $\boldsymbol D_{up}$ into degradation representations $\boldsymbol {\tilde{D}}$  and $\boldsymbol D $, where $\boldsymbol {\tilde{D}}=f_{rb} ( \boldsymbol D_{up}) $ and $\boldsymbol D=E_{d}(\boldsymbol {\tilde{D}})$.

Next, inspired by the Mixture-of-Experts~\cite{cao2023multi, shazeer2017outrageously, he2024frequency}, we construct a router to dynamically allocate the degradation representation $\boldsymbol D $ to degradation regularization, thereby adaptively selecting degradation kernel generators of different scales. The learned router $\boldsymbol {\mathcal{R}}$ is formulated as:
\begin{equation}
   \boldsymbol {\mathcal{R}} = \sigma  (  topK   (  E_{r}  (  \boldsymbol D_{up}   )     )    ) ,
\end{equation}
where $\sigma$ and $E_{r}$ are the softmax function and the routing encoder, respectively. $topK$ indicates the adaptive allocation of $\boldsymbol D $ to the top $k$ degradation kernel generators from $g$ candidate generators based on their scores.

\begin{figure}[t]
\centering
\includegraphics[width=1\columnwidth]{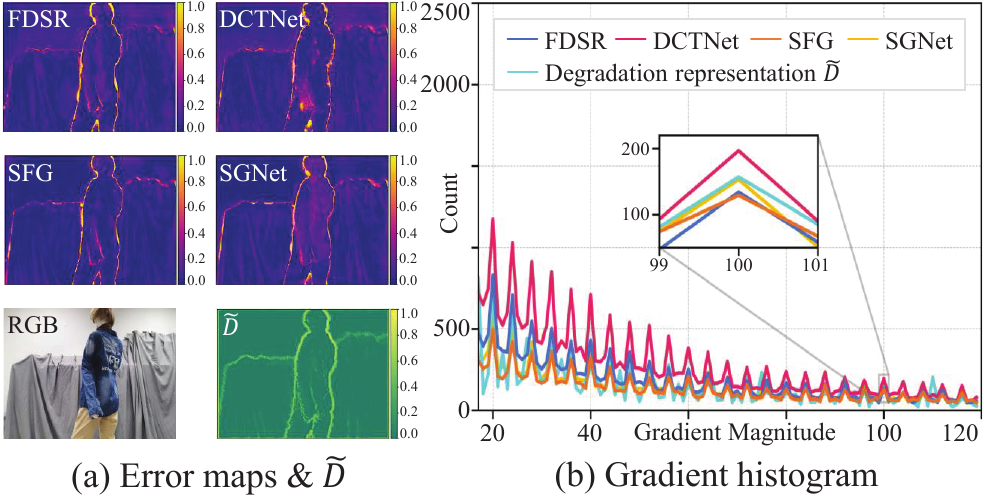}\\
\vspace{-4pt}
\caption{Visualization of error maps and degradation representation $\boldsymbol {\tilde{D}}$ (a), and their gradient histograms (b).}\label{fig:error_D}
\vspace{-3pt}
\end{figure}

\noindent{\textbf{Degradation Regularization.}} As depicted in Fig.~\ref{fig:pipeline} (upper right), given $\boldsymbol D $ as input, we first select $k$ degradation kernel generators of different scales under the assignment of router $\boldsymbol {\mathcal{R}}$, adaptively producing a multi-scale degradation kernel set $\mathbb{S} $. As an example, the degradation kernel $\boldsymbol {s}_{2i+1}$ of size $(2i+1)\times (2i+1)$ in $\mathbb{S} $ is represented as:
\begin{equation}
   \boldsymbol {s}_{2i+1}  = f_{g}^{2i+1} ( \boldsymbol {\mathcal{R}}, \boldsymbol D ), i\ge 1,
\end{equation}
where $f_{g}^{2i+1}$ refers to the degradation kernel generator with a size of $(2i+1)\times (2i+1)$, consisting of MLP. 

Then, the filtering and summation modules take the degradation kernel set $\mathbb{S} $ and the predicted HR depth $\boldsymbol D_{hr} $ as inputs to synthesize the degraded depth $\boldsymbol D_{d} $, which is used to supervise the learning of $\boldsymbol {\tilde{D}}$  and $\boldsymbol D $. Specifically, each degradation kernel in $\mathbb{S} $ is employed as a convolution kernel to individually convolve with $\boldsymbol D_{hr} $. The resulting convolution outputs are summed to generate $\boldsymbol D_{d} $:
\begin{equation}
   \boldsymbol D_{d} =\textstyle \sum_{j=1}^{k} \Lambda  ( \mathbb{S} _{j} , \boldsymbol D_{hr}),
\end{equation}
where $\Lambda $  represents the convolution operation.

\begin{figure}[t]
\centering
\includegraphics[width=0.95\columnwidth]{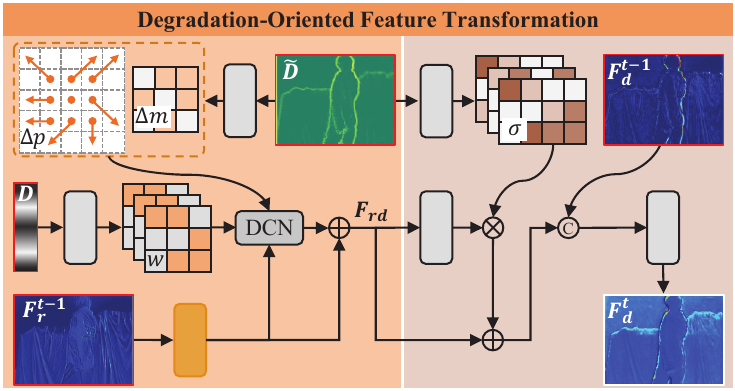}\\
\vspace{-2pt}
\caption{Details of DOFT. $\otimes$ is element-wise multiplication while \textcircled{c} is concatenation. Orange rectangular box: residual group~\cite{zhang2018image}.}\label{fig:doft}
\vspace{-3pt}
\end{figure}
\begin{table*}[t]
\footnotesize
	\centering
	\LARGE
	\resizebox{1\linewidth}{!}{
\begin{tabular}{c|ccccccccccccc}
\toprule 
RMSE		&DJF~\cite{li2016deep} &DJFR~\cite{li2019joint} &CUNet~\cite{deng2020deep} & DKN~\cite{kim2021deformable}  & FDKN~\cite{kim2021deformable}  & FDSR~\cite{he2021towards}   &DCTNet~\cite{zhao2022discrete}   &SUFT~\cite{shi2022symmetric}	  &SSDNet~\cite{zhao2023spherical}  & SFG~\cite{yuan2023structure}		&SGNet~\cite{wang2024sgnet}	&\textbf{DORNet-T}&\textbf{DORNet}\\
\midrule
Params. (M) 	&0.08     &0.08    &0.21    &1.16    &0.69    &0.60    &0.48    &22.01     &-       &63.53   &8.97             &0.46              &3.05 \\
RGB-D-D     &5.54     &5.52    &5.84    &5.08    &5.37    &5.49    &5.43    &5.41      &5.38    &3.88    &5.32             & \underline{3.84} & \textbf{3.42}	\\
TOFDSR     	&5.84     &5.72    &6.04    &5.50    &5.77    &5.03    &5.16 	 &4.37     &-       &4.52	 &\underline{4.33} &4.87	          &\textbf{4.21} \\
\bottomrule
\end{tabular}}
\vspace{-8pt}
\caption{Quantitative comparison with existing state-of-the-art methods on the real-world RGB-D-D and TOFDSR datasets.}\label{tab:NoNoisy}
\vspace{-8pt}
\end{table*}

\begin{table*}[t]
\footnotesize
	\centering
	\large
	\resizebox{1\linewidth}{!}{
\begin{tabular}{c|cccccccccccc}
\toprule 
RMSE		&DJF~\cite{li2016deep} &DJFR~\cite{li2019joint} &CUNet~\cite{deng2020deep} & DKN~\cite{kim2021deformable}  & FDKN~\cite{kim2021deformable}  & FDSR~\cite{he2021towards}   &DCTNet~\cite{zhao2022discrete}   &SUFT~\cite{shi2022symmetric}	   			& SFG~\cite{yuan2023structure}  &SGNet~\cite{wang2024sgnet} 				&\textbf{DORNet-T}&\textbf{DORNet}\\
\midrule
RGB-D-D  	&5.83   &5.78   &5.96   &5.52   &5.69   &5.66  	&5.61 	&5.53  		& \underline{4.08}	&5.44   &4.24	            & \textbf{3.68}		\\
TOFDSR      &8.21   &7.03   &8.64   &5.96   &6.86   &5.58  	&5.46 	&5.08       &5.46               &5.11   &\underline{5.07}	&\textbf{4.47} \\
\bottomrule
\end{tabular}}
\vspace{-8pt}
\caption{Quantitative comparison of joint DSR and denoising on the real-world RGB-D-D and TOFDSR datasets.}\label{tab:NoisyLR}
\vspace{-8pt}
\end{table*}

Next, we introduce a pre-trained VGG19~\cite{simonyan2014very} to map $\boldsymbol D_{hr} $, $\boldsymbol D_{d} $, and $\boldsymbol D_{up}$ to the latent space, yielding negative sample $\boldsymbol F_{n} $, anchor sample $\boldsymbol F_{a} $, and positive sample $\boldsymbol F_{p} $, respectively. These samples are used as inputs for the contrastive loss $ \mathcal{L}_{cont} $,  pulling the degraded depth $\boldsymbol D_{d} $ closer to the LR depth $\boldsymbol D_{up}$ and pushing it away from the HR depth $\boldsymbol D_{hr} $, thereby facilitating the learning of degradation representations:
\begin{equation}\label{eq:Lcont}
   \mathcal{L}_{cont}=\textstyle \sum_{z=1}^{m} \alpha _{z} \cdot \frac{ f_{l1} ( \boldsymbol F_{p}^{z}-\boldsymbol F_{a}^{z})  }{ f_{l1} ( \boldsymbol F_{n}^{z}-\boldsymbol F_{a}^{z}) } ,
\end{equation}
where $m$ denotes the number of latent space features, and $\alpha$ is a weight vector. $f_{l1}$ refers to the $L_{1} $ distance. 

Additionally, a degradation loss $\mathcal{L} _{deg}$ is employed to further optimize the degradation learning:
\begin{equation}\label{eq:Ldeg}
   \mathcal{L} _{deg}=\frac{1}{Q} \textstyle  \sum_{q=1}^{Q}    \| \boldsymbol D_{up}^{q} - \boldsymbol D_{d}^{q} \| _{1},
\end{equation}
where $Q$ refers to the number of training samples. $ \| \cdot  \| _{1} $ represents the $L_{1} $ loss function.

Fig.~\ref{fig:error_D} presents a visual comparison of the learned degradation representation $\boldsymbol {\tilde{D}}$ with the error maps of previous methods, as well as a comparison of their gradient histograms. The visualizations and gradient distributions demonstrate that $\boldsymbol {\tilde{D}}$ successfully learns the degraded depth structures that is challenging for previous approaches to recover, thereby providing targeted guidance for enhancing these severely degraded depth features. 

More importantly, degradation regularization is only applied in the training to facilitate the learning of degradation representations, and it does not introduce any additional computational overhead during testing.

\begin{figure}[t]
\centering
\includegraphics[width=0.93\columnwidth]{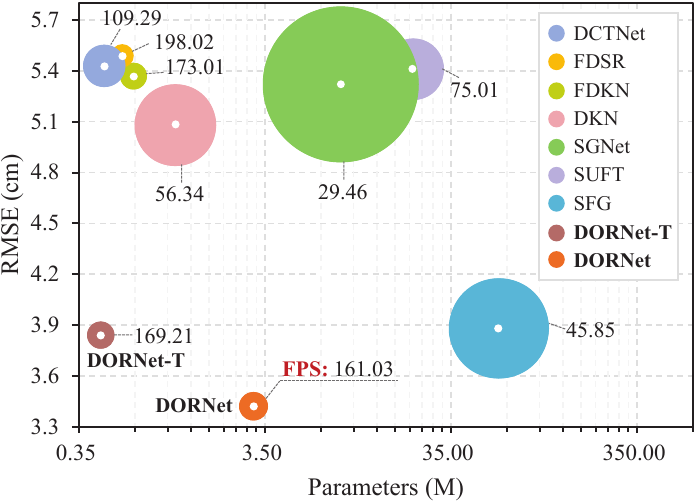}\\
\vspace{-5pt}
\caption{Complexity on RGB-D-D (w/o Noisy) tested by a 4090 GPU. A larger circle diameter indicates a higher inference time.}\label{fig:params}
\vspace{-5pt}
\end{figure}

\subsection{Degradation-Oriented Fusion}
As shown in the orange part of Fig.~\ref{fig:pipeline}, $\boldsymbol D_{lr}$ is first input into bicubic upsampling. Then, the upsampled LR depth and $\boldsymbol I$ are mapped to $\boldsymbol F_{d}^{0}$ and $\boldsymbol F_{r}^{0}$, respectively. 

Next, we take $\boldsymbol F_{d}^{0}$, $\boldsymbol F_{r}^{0}$, $\boldsymbol {\tilde{D}}$,  and $\boldsymbol D $ as inputs and recursively conduct multiple DOFT to selectively propagate RGB content into the depth features, generating the enhanced depth feature $\boldsymbol F_{d}^{t}$:
\begin{equation}
   \boldsymbol F_{d}^{t}=f_{do}^{t}   (  \boldsymbol {\tilde{D}},\boldsymbol D,\boldsymbol F_{d}^{t-1},\boldsymbol F_{r}^{t-1}  )  ,
\end{equation}
where $f_{do}^{t}$ refers to $t$-th DOFT.

Finally, the HR depth $\boldsymbol D_{hr}$ is predicted by fusing depth features $\boldsymbol F_{d}^{0}$ and $\boldsymbol F_{d}^{t}$:
\begin{equation}
   \boldsymbol D_{hr}=f_{c} ( \boldsymbol F_{d}^{0} + f_{c} ( \boldsymbol F_{d}^{t}  )) ,
\end{equation}
where $f_{c}$ refers to the convolutional layer, indicated by the gray rectangular box in Fig.~\ref{fig:pipeline} and~\ref{fig:doft}.

\begin{figure}[t]
\centering
\includegraphics[width=0.88\columnwidth]{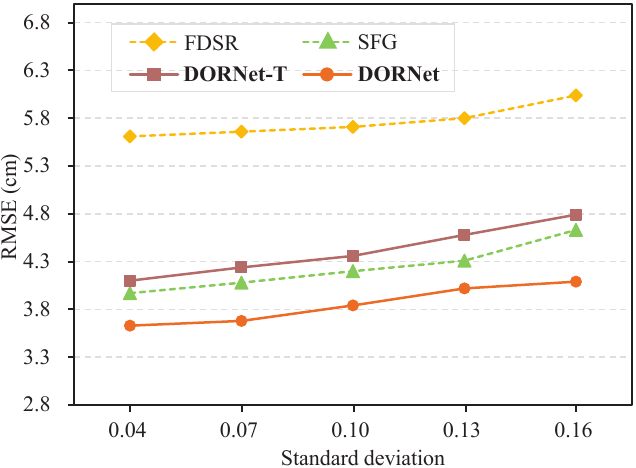}\\
\vspace{-5pt}
\caption{Robustness with different noises on RGB-D-D.}\label{fig:NoiseRob}
\vspace{-5pt}
\end{figure}

\noindent{\textbf{Degradation-Oriented Feature Transformation.}} Fig.~\ref{fig:doft} shows that DOFT includes degradation-oriented RGB feature learning (left part) and RGB-D feature fusion (right part). Specifically, DOFT first maps $\boldsymbol {\tilde{D}}$ to the offset $\triangle p$ and modulation scalar $\triangle m$, both of which are utilized to dynamically adjust the receptive field of the deformable convolution (DCN)~\cite{zhu2019deformable}  $f_{d}$. Then, we generate the weights $w$ of the DCN using $\boldsymbol D $ to focus its attention on RGB features that match the degraded depth structures.


Next, given RGB feature $\boldsymbol F_{r}^{t-1}$ as input, $\triangle p$, $\triangle m$, and $w$ are together used to adaptively learn the RGB feature $\boldsymbol F_{rd}$ aligned with the degradation representations:
\begin{equation}
    \boldsymbol F_{rd}=f_{d} ( f_{rg}  ( \boldsymbol F_{r}^{t-1} ), \bigtriangleup p,\bigtriangleup m,w)+f_{rg} ( \boldsymbol F_{r}^{t-1} ),
\end{equation}
where $f_{rg}$ is the residual group~\cite{zhang2018image}, a feature extraction unit consisting of residual block and channel attention.

Finally, we encode $\boldsymbol {\tilde{D}}$ as an affinity coefficient $\sigma $ for the selective transfer of learned RGB feature $ \boldsymbol F_{rd}$ to the depth, resulting in the enhanced depth feature $ \boldsymbol F_{d}^{t}$:
\begin{equation}
    \boldsymbol F_{d}^{t}=f_{c}   (   [ \boldsymbol F_{d}^{t-1},\sigma \otimes f_{c}(\boldsymbol F_{rd}) + \boldsymbol F_{rd} ]   )  ,
\end{equation}
where $\boldsymbol F_{d}^{t-1}$ is the input depth feature of DOFT. $ [ \cdot ] $ denotes concatenation. $\otimes$ refers to element-wise multiplication.

\begin{figure*}[t]
\centering
\includegraphics[width=2\columnwidth]{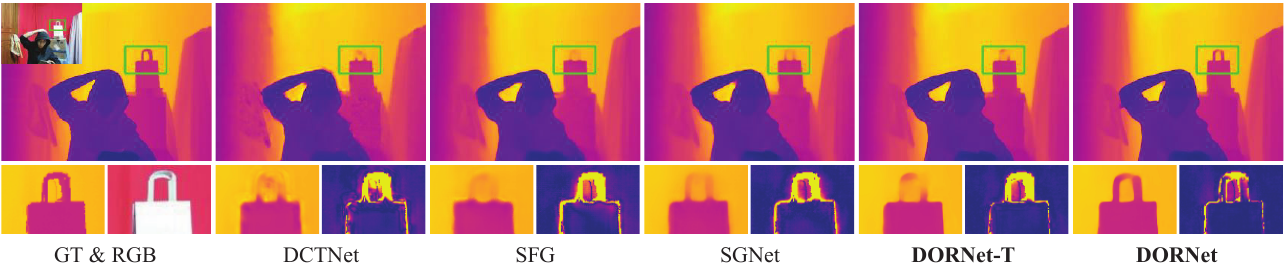}\\
\vspace{-8pt}
\caption{Visual results (left) and error maps (right) on the real-world RGB-D-D dataset (w/o Noise).}\label{fig:rgbdd_filled}
\vspace{-8pt}
\end{figure*}

\begin{figure*}[t]
\centering
\includegraphics[width=2\columnwidth]{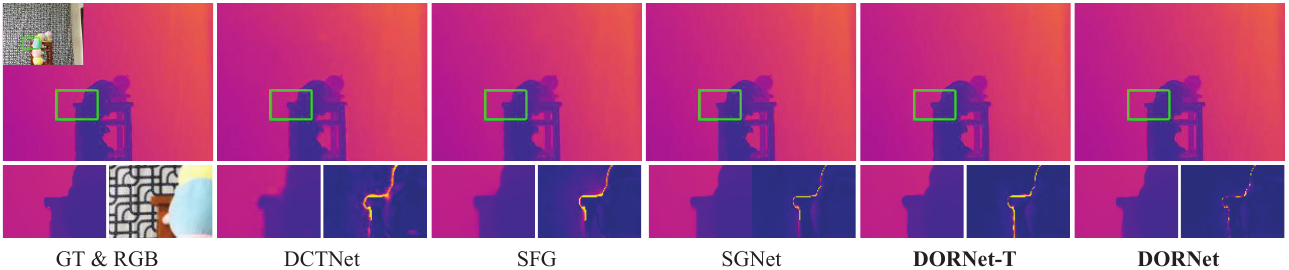}\\
\vspace{-8pt}
\caption{Visual results (left) and error maps (right) on the real-world TOFDSR dataset  (w/o Noise).}\label{fig:tofdsr_filled}
\vspace{-8pt}
\end{figure*}

\subsection{Loss Function}
Given the predicted HR depth $\boldsymbol D_{hr}$ and the ground-truth depth $\boldsymbol D_{gt}$, we first introduce the reconstruction loss $\mathcal{L} _{rec}$ to optimize our DORNet:
\begin{equation}
   \mathcal{L} _{rec}=\frac{1}{Q} \textstyle \sum_{q=1}^{Q}  \| \boldsymbol D_{gt}^{q} - \boldsymbol D_{hr}^{q} \| _{1}.
\end{equation}

Then, combining Eqs.~\eqref{eq:Lcont} and \eqref{eq:Ldeg}, the total training loss $\mathcal{L} _{total}$ is defined as:
\begin{equation}
   \mathcal{L} _{total}=\mathcal{L} _{rec} +\lambda _{1}\mathcal{L} _{deg} +\lambda _{2}\mathcal{L} _{cont},
\end{equation}
where $\lambda _{1}$ and $\lambda _{2}$ are hyper-parameters.



\section{Experiments}
\label{sec:exp}

\subsection{Experimental Setups}
\noindent{\textbf{Datasets.}} We conduct extensive experiments on both real-world RGB-D-D~\cite{he2021towards}, TOFDSR~\cite{yan2024tri}, and synthetic NYU-v2~\cite{silberman2012indoor} datasets. Specifically, for the RGB-D-D, the training set comprises $2,215$ RGB-D pairs, while the test set contains $405$ pairs. Additionally, the colorization method~\cite{levin2004colorization} is used to fill in the raw LR depth of the TOFDC~\cite{yan2024tri}, obtaining the TOFDSR that includes $10K$ RGB-D pairs for training and $560$ pairs for testing. In the real-world scenarios, the LR depth is obtained using the ToF camera of the Huawei P30 Pro. Following~\cite{kim2021deformable, zhao2022discrete, wang2024sgnet}, the synthetic NYU-v2 consists of $1,000$ RGB-D pairs for training and $449$ pairs for testing, with the LR depth generated by bicubic downsampling from the GT depth.

To weaken the interference of erroneous depth in the TOFDSR dataset, all methods calculate the loss and RMSE only for valid pixels where the GT depth is within the range of $0.1m$ to $5m$. For the RGB-D-D and NYU-v2 datasets, we maintain the same settings as in previous methods~\cite{he2021towards, zhao2022discrete, he2021towards}.

\begin{figure*}[t]
\centering
\includegraphics[width=2\columnwidth]{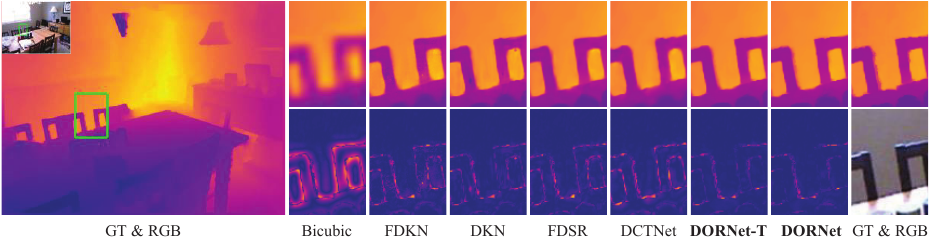}\\
\vspace{-6pt}
\caption{Visual results (top) and error maps (bottom) on the synthetic NYU-v2 dataset ($\times 8$).}\label{fig:nyu}
\vspace{-6pt}
\end{figure*}

\begin{table*}[t]
\footnotesize
	\centering
	\huge
	\resizebox{1\linewidth}{!}{
\begin{tabular}{c|ccccccccccccc}
\toprule
RMSE		&PAC~\cite{su2019pixel}   &CUNet~\cite{deng2020deep} &DKN~\cite{kim2021deformable} & FDSR~\cite{he2021towards}  & GraphSR~\cite{de2022learning}  & DCTNet~\cite{zhao2022discrete}   &SUFT~\cite{shi2022symmetric}   &DADA~\cite{metzger2023guided}    	&SSDNet~\cite{zhao2023spherical}	 	&SFG~\cite{yuan2023structure}	&SGNet~\cite{wang2024sgnet}	&\textbf{DORNet-T}&\textbf{DORNet}\\
\midrule
Params. (M)    &-      &0.21  &1.16  &0.60  &32.53  &0.48  &22.01              &32.53            &-     &63.53  &35.42          &0.46    &3.05	\\ 
$\times 4$    &1.89   &1.92  &1.62  &1.61  &1.79   &1.59  &\underline{1.12}   &1.54             &1.60  &1.45   &\textbf{1.10}  &1.33	&1.19	\\
$\times 8$    &3.33   &3.70  &3.26  &3.18  &3.17   &3.16  &\underline{2.51}   &2.74             &3.14  &2.84   &\textbf{2.44}  &2.90	&2.70	\\
$\times 16$   &6.78   &6.78  &6.51  &5.86  &6.02   &5.84  &4.86 			  &\underline{4.80} &5.86  &5.56   &\textbf{4.77}  &5.95	&5.60	\\
\bottomrule
\end{tabular}}
\vspace{-6pt}
\caption{Quantitative comparison with existing  state-of-the-art methods on the synthetic NYU-v2 dataset.}\label{tab:NYU}
\vspace{-8pt}
\end{table*}

\noindent{\textbf{Implementation Details.}} We employ the root mean square error (RMSE) in centimeter as the evaluation metric to be consistent with previous DSR methods~\cite{sun2021learning, he2021towards, yuan2023structure, zhao2023spherical}. The Adam~\cite{kingma2014adam} optimizer with an initial learning rate of $1\times 10^{-4} $ is used to train our DORNet. Besides, we implement our model in PyTorch using the NVIDIA GeForce RTX 4090. The hyper-parameters are set as $\lambda _{1} = \lambda _{2} = 0.1$. 

\subsection{Comparison with the State-of-the-Art}
We compare DORNet with popular methods, \emph{i.e.}, DJF~\cite{li2016deep}, DJFR~\cite{li2019joint}, PAC~\cite{su2019pixel}, CUNet~\cite{deng2020deep}, DKN~\cite{kim2021deformable}, FDKN~\cite{kim2021deformable}, FDSR~\cite{he2021towards}, GraphSR~\cite{de2022learning}, DCTNet~\cite{zhao2022discrete}, SUFT~\cite{shi2022symmetric}, DADA~\cite{metzger2023guided}, SSDNet~\cite{zhao2023spherical}, SFG~\cite{yuan2023structure}, and SGNet~\cite{wang2024sgnet}. To ensure a fair comparison, we directly cite the data from their papers for methods with existing experimental results. For other approaches, we utilize their released code to retrain and test under the same settings.

\noindent{\textbf{Comparison on Real-World Dataset.}} Tab.~\ref{tab:NoNoisy} indicates that our DORNet outperforms other advanced methods on the real-world RGB-D-D and TOFDSR datasets. From the first two rows of Tab.~\ref{tab:NoNoisy}, it can be seen that DORNet surpasses SFG~\cite{yuan2023structure} by $0.46cm$ on RGB-D-D while also significantly reducing the number of parameters. Moreover, the third row demonstrates that our method decreases the RMSE by $0.12cm$ on TOFDSR compared to SGNet~\cite{wang2024sgnet}.

Furthermore, Figs.~\ref{fig:rgbdd_filled} and \ref{fig:tofdsr_filled} present the visual results on the RGB-D-D and TOFDSR. In the error maps, a brighter color means a larger error. Obviously, for severely degraded LR depth, our method succeeds in recovering accurate depth structures. For instance, the handbag in Fig.~\ref{fig:rgbdd_filled} predicted by our method is more precise than others. Additionally, the error maps in Fig.~\ref{fig:tofdsr_filled} show that DORNet reconstructs HR depth with fewer errors.

Fig.~\ref{fig:params} illustrates that our method achieves a satisfactory balance among parameters, inference time, FPS, and performance. For example, compared to lightweight DCTNet ($0.48M$), our DORNet-T ($0.46M$) reduces RMSE by $29\%$ and inference time by $35\%$. Moreover, DORNet surpasses the second-best approach by $11\%$ while significantly decreasing both parameters and inference time.

\noindent{\textbf{Robustness to Noise.}} Tab.~\ref{tab:NoisyLR} demonstrates that our method exhibits robustness in noisy environments. Similar to previous approaches~\cite{kim2021deformable, yuan2023structure}, we add Gaussian noise (mean $0$ and standard deviation $0.07$) and Gaussian blur (standard deviation $3.6$) to upsampled LR depth as new input. We can see that DORNet outperforms SFG~\cite{yuan2023structure} by $0.40cm$ in RMSE on the RGB-D-D. For experiments on adding noise before LR depth pre-upsampling, please see our appendix.

Fig.~\ref{fig:NoiseRob} shows the comparison across different noise levels, with the standard deviation of Gaussian noise ranging from $0.04$ to $0.16$, while the Gaussian blur remains unchanged. We can observe that as the noise levels increase, the performance of all methods gradually declines. However, our DORNet consistently outperforms other approaches at each noise level. For instance, our method reduces RMSE by $0.36cm$ (standard deviation $0.10$) and by $0.29cm$ (standard deviation $0.13$)  compared to SFG \cite{yuan2023structure}.


\noindent{\textbf{Comparison on Synthetic Dataset.}} Tab.~\ref{tab:NYU} shows that our method achieves comparable performance on the NYU-v2 dataset. The first row lists the model parameters with a scale factor of 4. For example, compared to the SGNet~\cite{wang2024sgnet}, our DORNet significantly reduces the parameters by $91\%$, while only increasing the RMSE by $8\%$ ($\times 4$). Furthermore, for lightweight DSR, our DORNet-T outperforms DCTNet by $16\%$ and FDSR by $17\%$ in RMSE ($\times 4$). Fig.~\ref{fig:nyu} reveals that the depth structures predicted by our method is more closely aligned with the ground-truth depth. For instance, the edges of chair exhibit less error than others.

In summary, all of these quantitative comparisons and visual results demonstrate that our method effectively enhances the performance of real-world DSR. 

\subsection{Generalization Ability}
To further evaluate the generalization ability of our method, we implement it on \textbf{pan-sharpening} and \textbf{depth completion} tasks. Please see our appendix for the details. 

\begin{table}[t]
	\centering
 \small
	    \resizebox{0.98\linewidth}{!}{ 
		\begin{tabular}{l|ccc}
			\toprule 
				Methods   	 			 	           &Params (M)       &w/o Noisy 	 &w/ Noisy    	\\ 
			\midrule
				baseline + DASR~\cite{wang2021unsupervised}  &3.44             &3.86 	         &4.06         \\
                baseline + KDSR~\cite{xia2023knowledge}      &3.73             &3.65	         &3.86         \\
				baseline + DL \& DR (\textbf{Ours})          &\textbf{3.05}    &\textbf{3.42}  &\textbf{3.69} \\
			\bottomrule  
		\end{tabular}}
  \vspace{-5pt}
		\caption{Comparison of different degradation learning methods on the real-world RGB-D-D dataset. DL indicates Degradation Learning, while DR refers to Degradation Regularization.} \label{tab:DL_DR}
  \vspace{-6pt}
\end{table}

\begin{figure*}[t]
\centering
\includegraphics[width=1.95\columnwidth]{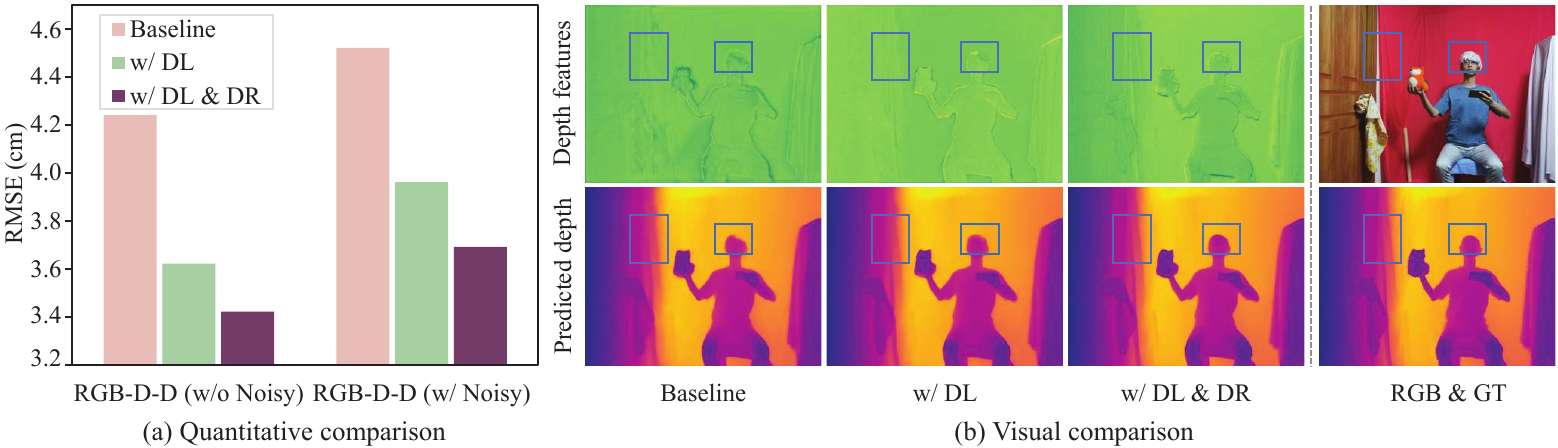}\\
\vspace{-7pt}
\caption{Ablation study of degradation learning (DL) and degradation regularization (DR) on the RGB-D-D dataset.}\label{fig:abl_dl_dr}
\vspace{-5pt}
\end{figure*}

\begin{figure*}[t]
\centering
\includegraphics[width=1.95\columnwidth]{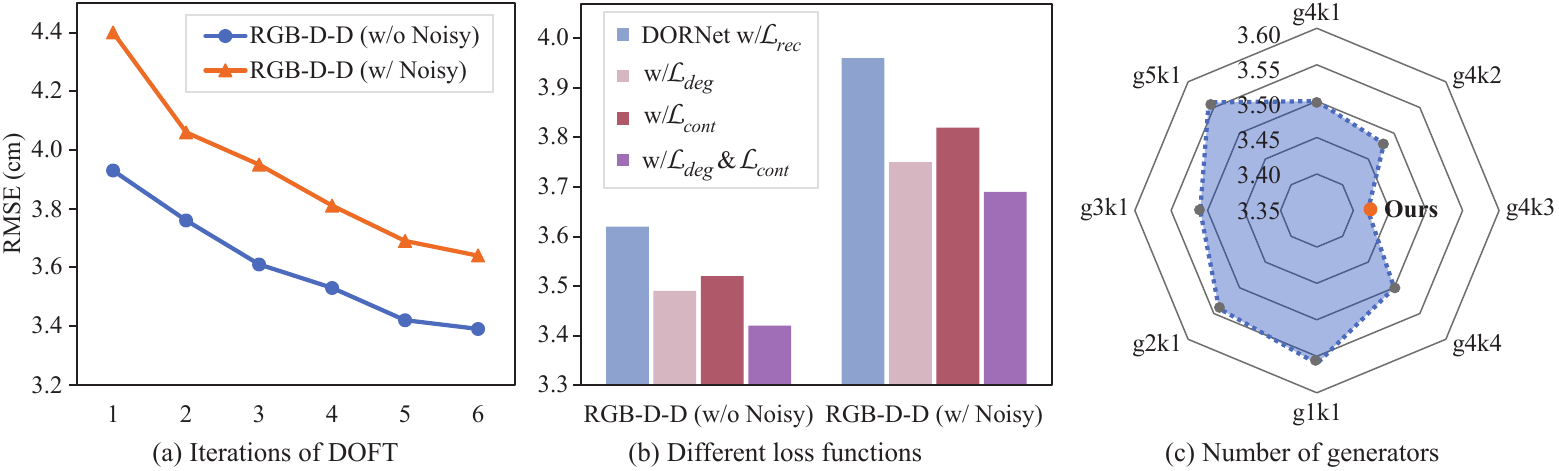}\\
\vspace{-7pt}
\caption{Ablation study of DORNet with (a) different numbers of DOFT, (b) different loss functions, and (c) different numbers of degradation kernel generators. `g4k3': DR selects 3 (k) out of 4 generators (g) of size $  (  2i+1  )  \times   (  2i+1  )  $, $1\le i\le 4$, based on the router.}\label{fig:abl_dlr}
\vspace{-8pt}
\end{figure*}

\subsection{Ablation Studies}
\noindent{\textbf{Degradation Learning and Regularization.}} 
Fig.~\ref{fig:abl_dl_dr} and Tab.~\ref{tab:DL_DR} present the ablation study of degradation learning (DL) and degradation regularization (DR). For the baseline, we first remove the entire DL and DR in DORNet. Then, we utilize concatenation to replace all DOFT. Additionally, only the reconstruction loss  is used during the training. 

Fig.~\ref{fig:abl_dl_dr}(a) reveals that DL significantly reduces RMSE by modeling the degradation representations. When DR is combined, our method achieves the best performance. For example, DORNet outperforms the baseline by $0.82cm$ (w/o Noisy) and $0.83cm$ (w/ Noisy). Fig.~\ref{fig:abl_dl_dr}(b) presents the visual results of the depth features and predicted depth. Compared to the baseline, DL contributes to generating clearer structures. When DR is employed together with DL, our approach produces more accurate depth.

Furthermore, Tab.~\ref{tab:DL_DR} lists the comparison results of DL and DR with previous degradation learning methods. Specifically, we replace the entire DL and DR with the degradation learning modules from DASR~\cite{wang2021unsupervised} and KDSR~\cite{xia2023knowledge}, respectively. It can be observed that our approach surpasses DASR by $0.44cm$ and KDSR by $0.23cm$ in RMSE (w/ Noisy). These results further demonstrate that our DL and DR can learn more accurate degradation representations and effectively enhance DSR performance. 

\noindent{\textbf{Different Recursion Numbers of DOFT.}} Fig.~\ref{fig:abl_dlr}(a) depicts the ablation study of different iterations of DOFT. The baseline is the entire DORNet with all loss functions. It is evident that performance incrementally improves as the number of DOFT iterations increases. When the number of iterations reaches $6$, the reduction in RMSE begins to slow down. To better trade-off between the model complexity and performance, our DORNet iterates $5$ DOFT.

\noindent{\textbf{Different Loss Functions.}} Fig.~\ref{fig:abl_dlr}(b) presents the ablation study of different loss functions. The baseline is the entire DORNet using only the reconstruction loss $\mathcal{L} _{rec}$. Obviously, we can see that both the degradation loss $\mathcal{L} _{deg}$ and contrastive loss $\mathcal{L} _{cont}$ contribute to performance improvement. When $\mathcal{L} _{deg}$ and $\mathcal{L} _{cont}$ are deployed together, our method achieves the lowest RMSE. For example, compared to the baseline, our DORNet decreases the RMSE by $0.20cm$ (w/o Noisy) and $0.27cm$ (w/ Noisy) on RGB-D-D.

\noindent{\textbf{Number of Generators.}} Fig.~\ref{fig:abl_dlr}(c) shows the ablation study of different numbers of degradation kernel generators on the RGB-D-D dataset (w/o Noisy). The baseline is the entire DORNet with $\mathcal{L} _{rec}$, $\mathcal{L} _{deg}$, and $\mathcal{L} _{cont}$.  We conduct experiments with $8$ sets of different generator selection settings. As an example, `g4k3' indicates that DR adaptively selects $3$ out of $4$ different-scale degradation kernel generators based on the router $\boldsymbol {\mathcal{R}}$, producing $3$ degradation kernels of different scales. Firstly, we observe that the RMSE of `g4k1' is lower than that of `g1k1', `g2k1', `g3k1', and `g5k1', indicating that more generators may not necessarily result in better performance. Secondly, `g4k3' achieves better DSR performance than `g4k1', `g4k2', and `g4k4'. Therefore, we select `g4k3' as the setting for DORNet.

\section{Conclusion}
\label{sec:con}
In this paper, we proposed the degradation oriented and regularized network, a novel real-world DSR solution that learns degradation representations of low-resolution depth to provide targeted guidance. 
Specifically, we designed a self-supervised degradation learning strategy to model the degradation representations using routing selection-based degradation regularization. This enables label-free implicit degradation learning that adaptively addresses unknown degradation in real-world scenes. 
Furthermore, we developed a degradation-oriented feature transformation module to perform effective RGB-D fusion. Based on the learned degradation priors, the module selectively propagates RGB content into depth, thereby restoring accurate high-resolution depth. 
Extensive experiments demonstrate the effectiveness and superiority of our method. 

\section*{Acknowledgements}
This work was supported by the National Science Fund of China under Grant Nos. U24A20330 and 62361166670.